\documentclass[10pt, a4paper]{article}
\usepackage{lrec}
\usepackage{authblk}
\usepackage{multibib}
\usepackage{graphicx}
\usepackage{tabularx}
\usepackage{soul}
\usepackage{epstopdf}
\usepackage[latin1]{inputenc}
\usepackage{xstring}
\usepackage{pdfpages}
\usepackage{tabularx}
\newcolumntype{b}{X}
\newcolumntype{s}{>{\hsize=.5\hsize}X}

\usepackage{makecell}
\usepackage{diagbox}
\usepackage{multirow}
\usepackage{pdfpages}
\usepackage[ruled,linesnumbered,vlined]{algorithm2e}

\newcommand{\fn}[1]{\footnote{#1}}

\usepackage{caption}
\usepackage{subcaption}
\usepackage{diagbox}

\usepackage{verbatim}

\newcommand{\citet}[1]{\newcite{#1}}
\newcommand{\citep}[1]{\cite{#1}}

\newcommand{\Sec}[1]{Section~\ref{sec:#1}}
\newcommand{\Fig}[1]{Figure~\ref{fig:#1}}
\newcommand{\Tab}[1]{Table~\ref{tab:#1}}
\newcommand{\Algo}[1]{Algorithm~\ref{algo:#1}}
\newcommand{\Eq}[1]{Eq.~(\ref{eq:#1})}
\newcommand{\ra}{{$\rightarrow$}}
\newcommand{\task}[2]{\mbox{#1{\ra}#2}}

\usepackage{colortbl}

\usepackage{xcolor}
\definecolor{cR}{RGB}{255,204,204}
\definecolor{cG}{RGB}{204,255,204}
\definecolor{cB}{RGB}{204,204,255}
\definecolor{cK}{RGB}{221,221,221}
\definecolor{navyblue}{rgb}{0.0,0.0,0.5}

\newcommand{\bgR}[1]{\cellcolor{cR}#1}

\newcommand{\bgB}[1]{\cellcolor{cB}#1}
\newcommand{\bgK}[1]{\cellcolor{cK}#1}

\usepackage[hidelinks]{hyperref}
\hypersetup
{
    colorlinks=true,
    citecolor=navyblue,
    linkcolor=navyblue,
    filecolor=navyblue,
    urlcolor=navyblue,
}
\title
{
    Coursera Corpus Mining and Multistage Fine-Tuning \\ for Improving Lectures Translation
}

\name{Haiyue Song$^1$, Raj Dabre$^2$, Atsushi Fujita$^3$, Sadao Kurohashi$^4$}
\address
{
$^{1,4}$Kyoto University\\
$^{2,3}$National Institute of Information and Communications Technology \\
$^1$song@nlp.ist.i.kyoto-u.ac.jp, $^4$kuro@i.kyoto-u.ac.jp \\
$^{2,3}$\{raj.dabre, atsushi.fujita\}@nict.go.jp  \\
}

\begin{document}
\abstract
{
    Lectures translation is a case of spoken language translation and there is a lack of publicly available parallel corpora for this purpose. To address this, we examine a language independent framework for parallel corpus mining which is a quick and effective way to mine a parallel corpus from publicly available lectures at Coursera. Our approach determines sentence alignments, relying on machine translation and cosine similarity over continuous-space sentence representations. We also show how to use the resulting corpora in a multistage fine-tuning based domain adaptation for high-quality lectures translation. For Japanese--English lectures translation, we extracted parallel data of approximately 40,000 lines and created development and test sets through manual filtering for benchmarking translation performance. We demonstrate that the mined corpus greatly enhances the quality of translation when used in conjunction with out-of-domain parallel corpora via multistage training. This paper also suggests some guidelines to gather and clean corpora, mine parallel sentences, address noise in the mined data, and create high-quality evaluation splits. For the sake of reproducibility, we will release our code for parallel data creation.
    \newline
}

\maketitleabstract
\section{Introduction}
\label{sec:introduction}

In recent years, massive open online courses (MOOCs) have proliferated and have enabled people to attend lectures regardless of their geographical location. Typically, such lectures are taught by professors from certain universities and are made available through video recordings. It is common for these lectures to be taught in a particular language and the video is accompanied by subtitles. These subtitles are then translated into other languages so that speakers of those languages also benefit from the lectures. As manual translation is a time-consuming task and there are a large number of lectures, having a high-quality machine translation (MT) system can help ease distribute knowledge to a large number of people across the world.

Given the fact that most lectures are spoken in English, translating English subtitles to other languages is an urgent and important task.
On the other hand, there are also lectures taught in other languages than English. For instance, several universities in Japan offer online lecture courses, mostly taught in Japanese. Enabling non-Japanese speakers to participate in these courses by translating Japanese lecture subtitles into other languages, including English, is also an important challenge.
The TraMOOC project \citep{kordoni2016tramooc} aims at improving the accessibility of European languages through MT. They focus on collecting translations of lecture subtitles and constructing MT systems for eleven European and BRIC languages. However, the amount of parallel resources involving other languages, such as Chinese and Japanese, are still quite low.

Subtitle translation falls under spoken language translation. Past studies in spoken language translation mainly focused on subtitles for TED talks \citep{cettolo2012wit3}.  Even though the parallel data in this domain should be exploitable for lectures translation to some degree, university lectures are devoted mainly for educational purposes, and the subtle differences in domains may hinder translation quality.
To obtain high-quality parallel data, professional translators are typically employed to translate.
However, the cost is often very high and thus using this way to produce large quantities of parallel data is economically infeasible, especially for universities and non-profit organizations.
In the case of online lectures and talks, subtitles are often translated by crowdsourcing \citep{behnke2018improving} which involves non-professional translators. The resulting translation can thus be often inaccurate and quality control is indispensable.
There are many automatic ways to find parallel sentences from roughly parallel documents \citep{tiedemann2016finding}.\fn{cf. Comparable corpora, such as Wikipedia, i.e., pairs of documents containing the contents in same topic but their parallelism is not necessarily guaranteed and the corresponding sentences are not necessarily in the same order.}
 In particular, MT-based approaches are quite desirable because of their simplicity and it is possible to use existing translation models to extract additional parallel data. However, using an MT system trained on data from another domain can give unreliable translations which can lead to parallel data of low quality.

In this paper, we propose a new method which combines machine translation and similarities of sentence vector representations to automatically align sentences between roughly aligned document pairs. As we are interested in educational lectures translation, we focus on extracting parallel data from Coursera lectures.
Using our method, we have compiled a Japanese--English parallel corpus of approximately 40,000 lines. We have also created test and development sets, consisting of 2,000 and 500 sentences, respectively, through manual filtering. We will make the data publicly available so that other researchers can use it for benchmarking their MT systems. All our data splits are at a document level and thus can be used to evaluate techniques that exploit context. To show the value of our extracted parallel data, we conducted experiments using it in conjunction with out-of-domain corpora for Japanese--English lectures translation. We show that although the small in-domain corpus is ineffective by itself, it is very useful when combined with out-of-domain corpora using domain adaptation techniques.

The contributions of this paper are as follows.
\begin{itemize}\itemsep=0mm
	\item A simple but accurate technique to extract parallel sentences from noisy parallel document pairs.  We will make the code publicly available.\fn{\url{https://github.com/shyyhs/CourseraParallelCorpusMining}}
	\item A Japanese--English parallel corpus usable for benchmarking educational lectures translation: high quality development and test sets are guaranteed through manual verification, whereas the training data is automatically extracted and is potentially noisy.
	\item An extensive evaluation of robust domain adaptation techniques leveraging out-of-domain data to improve lectures translation for the Japanese--English pair.
\end{itemize}

\section{Related Work}

Our work focuses on three main topics: spoken language corpora in the educational domain, parallel corpora alignment, and domain adaptation for low-resource MT.

\subsection{Educational Spoken Language Corpora}

Most spoken language corpora come from subtitles of online videos and a sizeable portion of subtitles are available for educational videos. Such videos are recorded lectures that form a part of an online course provided by an organization which is usually non-profit. Nowadays, many MOOCs\fn{\url{http://mooc.org}} have become available and help people to conveniently acquire knowledge regardless of their location.
The TraMOOC project \citep{kordoni2016tramooc,kordoni2016enhancing} aims at providing access to multilingual subtitles of online courses by using MT. Coursera\fn{\url{https://www.coursera.org}} is an extremely popular platform for MOOCs and a large number of lectures have multilingual subtitles which are created by professional and non-professional translators alike. A similar MOOC site is Iversity.\fn{\url{https://iversity.org}}

Another existing spoken language corpora is for TED talks \citep{cettolo2012wit3}.\fn{\url{https://wit3.fbk.eu/mt.php?release=2017-01-ted-test}}  Most talks are for the purpose of educating people, even though they do not belong to the educational lectures domain. On a related note, Opensubtitles \citep{tiedemann2016finding}\fn{\url{https://www.opensubtitles.org}} is a collection of subtitles in multiple languages but mixes several domains.

\subsection{Parallel Corpus Alignment}

Extracting parallel data usable for MT involves crawling documents and aligning translation pairs in the corpora. To align translations, one can use crowdsourcing services \citep{behnke2018improving}. However, this can be extremely time-consuming if not expensive. Previous research \citep{abdelali2014amara} focused on collecting data from AMARA platform \citep{jansen2014amara}. They usually aim at European and BRIC languages, such as German, Polish, and Russian.

Using automatic alignment methods are more desirable, because they can help extract parallel sentences that are orders of magnitude larger than those that can be obtained by manual translation, including crowdsourcing. Although the quality of the extracted parallel sentences might be low, relying on comparable corpora can help address quality issues  \citep{wolk2015noisy} where one can use time-stamp to roughly align corresponding documents \citep{abdelali2014amara,tiedemann2016finding}. In order to obtain high-quality parallel data from these documents, MT-based methods \citep{sennrich2010mt,sennrich2011iterative,liu2018chinese} and similarity-based methods \citep{bouamor2018h2,wang2019target} can be combined with dynamic programming \citep{utsuro1994bilingual} for fast and accurate sentence alignment. The LASER tool \citep{chaudhary2019low}\fn{\url{https://github.com/facebookresearch/LASER}} offers another way to align sentence pairs automatically in an unsupervised fashion.

\subsection{Domain Adaptation for Neural Machine Translation}

At present, neural machine translation (NMT) is known to give higher quality of translation. To train a sequence-to-sequence model \citep{sutskever2014sequence}, attention-based model \citep{bahdanau2014neural} or self-attention based model \citep{vaswani2017attention}, we need a large parallel corpus for high-quality translation \citep{1604.02201,1706.03872}.
In the case of the news domain, there are many corpora, e.g., News Commentary \citep{tiedemann2012parallel}, containing large number of parallel sentences that enable high-quality translation.
In contrast, for educational lectures translation, only relatively small datasets are available.
Transfer learning through fine-tuning an out-of-domain model on the in-domain data \citep{luong2015stanford,sennrich2015improving,1604.02201,raj17} is the most common way to overcome the lack of data. However, approaches based on fine-tuning suffer from the problem of over-fitting which can be addressed by strong regularization techniques \citep{HinSal06,chelba2006adaptation,miceli-barone-etal-2017-regularization,thompson-etal-2019-overcoming}. Furthermore, the domain divergence between the out-of- and in-domain corpora is another issue.

\section{Our Framework for Mining Coursera Parallel Corpus}

\begin{figure}[t]
    \centering
    \includegraphics[scale=.45]{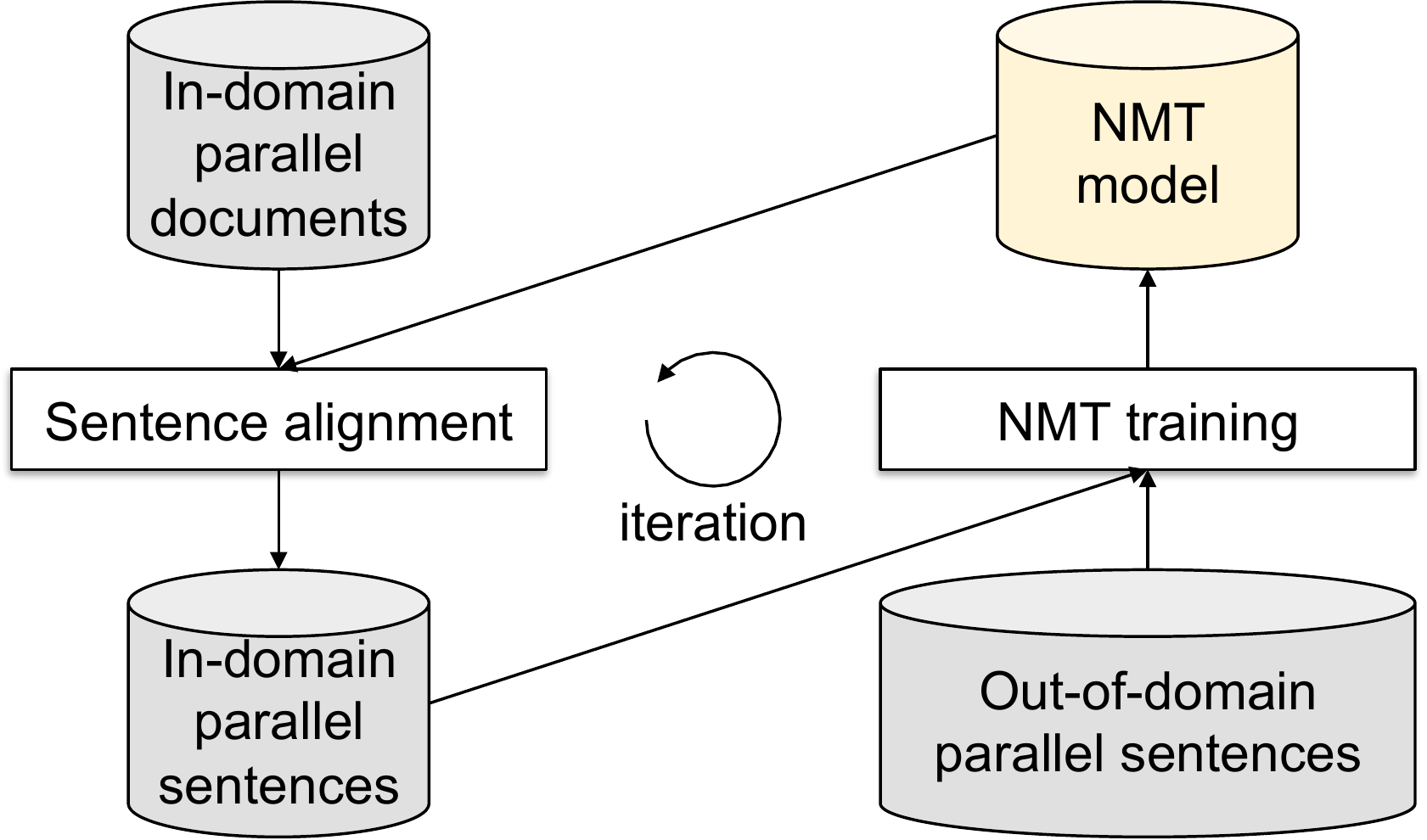}
    \caption{Overview of our framework.}
    \label{fig:framework}
\end{figure}

This section describes our general framework to compile a parallel corpus in educational lectures domain, relying on Coursera.
\Fig{framework} gives an overview of our framework, where we assume the availability of in-domain parallel documents (top-left), such as those available from Coursera, and out-of-domain parallel sentences (bottom-right).
We give details about the way we prepare the source document pairs, align the sentence pairs in the documents, and create evaluation splits for benchmarking.

\subsection{Crawling and Cleaning Parallel Documents}
\label{sec:crawlclean}

Our framework exploits a set of in-domain parallel documents, i.e., translated lectures subtitles, available at Coursera.
First, the list of available courses at Coursera is obtained, for instance by scraping.  Then, all the subtitles in all available languages are downloaded from each course in the list, for instance by using Coursera-dl.\fn{\url{https://github.com/coursera-dl/coursera-dl}}  Note that this results in a multilingual document-aligned subtitle corpus. From the extracted document pairs, we retain those in which the order of the sentences are roughly in the same order of the time-stamps of the lecture.

To obtain high-quality translations, the crawled parallel documents must be intensively cleaned.  We consider the following 5-step procedure.

\begin{description}
\item[Step 1. Normalizing Text Encoding:] First, all the documents are converted into UTF-8 and variants of character encodings were normalized (NFKC).
\item[Step 2. Detecting Language Mismatch:] The content of a document is sometimes of a different language than mentioned on the website.  Thus, we have to detect and exclude such mismatches.
  Language detection tools, such as langdetect \citep{raffel2019exploring},\fn{\url{https://github.com/Mimino666/langdetect}} and/or hand-written rules can be used.

\item[Step 3. Splitting Lines into Sentences:] Since not all lines within a document are segmented into sentences, sentence splitting is necessary. Punctuation marks can be regarded as the clue.  Files containing no punctuation marks are discarded, because we currently have no reliable way to deal with them.

\item[Step 4. Removing Meta Tokens:] Some tokens indicating meta-information, such as ``[Music]'' and ``$<<$,'' in each file are removed.

\item[Step 5. Eliminating Imbalanced Document Pairs:] Some document pairs are imbalanced in the sizes: one side has twice or more sentences than the other.  Such pairs are eliminated.
\end{description}

\subsection{Sentence Alignment}
\label{sec:alignment}

Given crawled and cleaned document pairs, we identify sentence alignments using dynamic programming (DP) as in \citet{utsuro1994bilingual},
assuming the monotonicity of subtitles: the corresponding sentences in each pair of documents are roughly in the same order.  Our assumption is based on the fact that the sentences in subtitle corpora are often constructed in accordance with the time-stamps of the sentences they correspond with. Consequently, comparing all pairs of sentences between document pairs is unnecessary.
Our DP algorithm relies on MT system, sentence similarity measure, and some constraints based on the nature of lectures subtitles.

\subsubsection{Training an Initial MT System}

To compute the similarity of arbitrary pair of sentences in two different languages, we first need to represent them in a common space.
One option is to translate one side into the other language using an MT system \citep{sennrich2010mt}.  To train such a system, we can leverage any existing parallel data in related or even distant domains.
The MT system should generate translations as accurately as possible.  In practice, domain adaptation techniques \citep{raj17} are most useful in training an accurate MT system.

\subsubsection{Similarity Measure}

The key component in the DP algorithm is the matching function, i.e., similarity measure in our context.  Existing methods, such as that in \citet{sennrich2010mt}, used sentence-level BLEU scores \citep{papineni:02:b} of machine-translated source sentence against the actual target language sentence as their similarity score: formally,
\begin{equation}
\mathit{Sim}_{\mathit{BLEU}}(f_{i},e_{j}) = \mathit{BLEU}(\mathit{MT}(f_{i}),e_{j}),
\label{eq:bleu}
\end{equation}
where $f_{i}$ and $e_{j}$ are the $i$-th sentence in the source document and the $j$-th sentence in the target document, respectively.
However, due to the lack of in-domain data, MT system can give only translations of low quality and thus the BLEU scores can be misleading, especially for distant language pairs, such as Japanese and English.

An alternative way is to directly compute cosine similarity of a given sentence pair \citep{bouamor2018h2}, relying on pre-trained multilingual word embeddings to represent sentences in different languages with the same vector space through element-wise addition of word embeddings \citep{mikolov2013distributed}.
However, cross-lingual embeddings are often not accurate for distant language pairs, especially if they have been pre-trained on data from another domain.

Taking inspiration from both these approaches, we employ MT combined with cosine similarity of sentence embeddings to measure the similarity of two sentences in different languages, formulated as follows.
\begin{equation}
\mathit{Sim}_{\mathit{EMB}}(f_{i},e_{j}) = \cos\left(\mathit{emb}(\mathit{MT}(f_{i})),\mathit{emb}(e_{j})\right).
\label{eq:emb}
\end{equation}
As in \Eq{bleu}, we first translate each sentence in the source language document into the target language. In practice, we prefer to have English as the target language, because this eliminates the need for cross-lingual vectors and this also enable us to use an abundance of high-quality English pre-trained vectors for several domains.
$\mathit{emb}(\cdot)$ represents the embedding of the given sentence, which can be computed by averaging the embeddings of words in that sentence, as in \citet{mikolov2013distributed}.

\subsubsection{Constraints}

To control the alignment quality, we should introduce the following three types of constraints in our DP algorithm.
\begin{itemize}\itemsep=0mm
\item A pair of sentences will not be a match if their similarity is lower than a pre-determined threshold, $\mathit{th}$.
\item A pair of sentences will not be a match if one of them is $k$ times longer than the other.
\item Only 1-1, 0-1, and 1-0 matching are allowed.
\end{itemize}

\subsection{Creating High-quality Evaluation Sets}
\label{sec:split}

To benchmark the performance of educational lectures translation, a high-quality test set is indispensable.  If we also have another set of high-quality translations, it can be a useful development set for tuning MT systems. We resort to manual cleaning of the scored and aligned sentence pairs obtained using the previous step.

We first sort all document pairs in the descending order of the average similarity of all aligned sentence pairs within each document pair. We then subject these sorted and sentence aligned pairs to human evaluation using \Algo{testdev} in order to obtain high-quality test and development sets, where the target volume of each set ($\mathit{volume}$) and document-level comparability ($\mathit{ratio}$) are the two parameters.  We use the remaining sentence aligned document pairs for training. Our test, development, and training sets are all constructed at the document level and thus our corpora can be used to evaluate document-level translation \citep{voita2019good,wang2019exploiting,tan-etal-2019-hierarchical}.

\begin{algorithm}[t]
\small
\SetKwInOut{Input}{Input}
\SetKwInOut{Output}{Output}
\Input{$\mathit{DocPairs}$, $\mathit{volume}$, $\mathit{ratio}$}
\Output{$\mathit{DocPairs}$, $\mathit{SentencePairs}$}
$\mathit{SentencePairs}\gets\{\}$\;
\While {$|\mathit{SentencePairs}|<\mathit{volume}$}
{
    $\mathit{DocPair}\gets$ pickBestDocPair($\mathit{DocPairs}$)\;
    $\mathit{DocPairs}\gets\mathit{DocPairs}\backslash\{\mathit{DocPair}\}$\;
    $\mathit{CandidatePairs}\gets$ getAlignments($\mathit{DocPair}$)\;
    $\mathit{Correct}\gets$ \{\}\;
    \ForEach{$\mathit{Pair}\in\mathit{CandidatePairs}$}
    {
        $\mathit{Judge}\gets$ manualEvaluation($\mathit{Pair}$)\;
        \If {$\mathit{Judge}$ == good}
        {
            $\mathit{Correct}\gets\mathit{Correct}\cup\{\mathit{Pair}\}$\;
        }
    }
    \If {$|\mathit{Correct}| > |\mathit{CandidatePairs}| * \mathit{ratio}$}
    {
        $\mathit{SentencePairs}\gets\mathit{SentencePairs}\cup\mathit{Correct}$\;
    }
}
\caption{Document-aware sentence filtering}
\label{algo:testdev}
\end{algorithm}

\begin{algorithm}[t]
\small
\SetKwInOut{Input}{Input}
\SetKwInOut{Output}{Output}
\Input{$\mathit{Doc}$, $n$, $m$, $\mathit{EnglishChar}$, $\mathit{JapaneseChar}$}
\Output{$\mathit{Label}$}
$\mathit{Sentences}\gets$pickRandomSentences($\mathit{Doc}$, $n$)\;
$\mathit{EnSentence}\gets 0$\;
$\mathit{JaSentence}\gets 0$\;
\ForEach{$\mathit{Sentence}\in\mathit{Sentences}$}
{
    $\mathit{EnChar}\gets 0$\;
    $\mathit{JaChar}\gets 0$\;
    $\mathit{OtherChar}\gets 0$\;
    \ForEach{$\mathit{Char}\in\mathit{Sentence}$}
    {
    \uIf{$\mathit{Char}\in\mathit{EnglishChar}$}
    {
        $\mathit{EnChar}\gets\mathit{EnChar}+1$\;
    }\uElseIf{$\mathit{Char}\in\mathit{JapaneseChar}$}
    {
        $\mathit{JaChar}\gets\mathit{JaChar}+1$\;
    }\Else{
        $\mathit{OtherChar}\gets\mathit{OtherChar}+1$\;
    }
    }
    \uIf{$\mathit{EnChar}>\mathit{JaChar}$}
    {
        $\mathit{EnSentence}\gets\mathit{EnSentence}+1$\;
    }\Else{
        $\mathit{JaSentence}\gets\mathit{JaSentence}+1$\;
    }
}
\uIf{$\mathit{EnSentence}\ge m$}
{
    $\mathit{Label}\gets\mathit{En}$\;
}\uElseIf{$\mathit{JaSentence}\ge m$}
{
    $\mathit{Label}\gets\mathit{Ja}$\;
}\Else
{
  $\mathit{Label}\gets\mathit{Noise}$\;
}
\caption{Language detection procedure}
\label{algo:handrules}

\end{algorithm}

\section{Creating Japanese--English Parallel Data}
\label{sec:jaen_data}

Although we have actually extracted document pairs for all available courses on Coursera,
henceforth, we report on an application of our framework to create a Japanese--English Coursera dataset.

\subsection{Cleaning Documents}

Our framework is mostly language independent. The only language specific processes are tokenization and language mismatch detection.
We first segmented the both English and Japanese paragraphs with full-stop (``.''), exclamation (``!''), and question marks (``?'') in Latin encoding and their full-width counterparts in UTF-8 followed by a space or the end of line.  Then, we tokenized Japanese and English sentences, using Juman++ \citep{jumanpp1}\fn{\url{https://github.com/ku-nlp/jumanpp}} and NLTK,\fn{\url{https://www.nltk.org}} respectively.

\Algo{handrules} shows our rule-based language detection procedure for the Japanese--English setting. It judges whether the given document is in Japanese or English according to the number of sentences within the document that belong to each language, where the language of each sentence is determined on the basis of the number of English and Japanese characters.  More specifically, we define a set of characters, $\mathit{EnglishChar}$, with ``a'' to ``z'' and ``A'' to ``Z,'' and another set of characters, $\mathit{JapaneseChar}$, with \emph{hiragana} and \emph{katakana}.  We also set the two thresholds: $n=10$ and $m=8$.

We evaluated the performance of the langdetect tool \citep{raffel2019exploring} and our algorithm on 100 sample documents, and found that the langdetect has one misclassification whereas ours worked perfectly with 1.00 precision and recall, presumably thanks to the cleanness of the Coursera data.
Considering that our simple method worked reasonably accurately, we chose the results of our method for the following steps.

\subsection{Creating Initial MT System}
As mentioned in \Sec{introduction}, the TED parallel corpus \citep{cettolo2012wit3} is from the spoken language domain and thus it is most similar to the spoken educational lectures domain. However, given its small size, it can lead to only an unreliable MT system.  Therefore, we decided to use a larger out-of-domain ASPEC corpus \citep{nakazawa2016aspec}\fn{We selected the best 1.0 million sentence pairs.} to build a better MT system. \Tab{aspected} gives the statistics of the ASPEC and TED corpora that we used to train our initial MT system. We compared fine-tuning and mixed fine-tuning approaches proposed by \citet{raj17}.  When performing mixed fine-tuning on the concatenation of both two corpora, the TED corpus was oversampled to match the size of the ASPEC corpus.
We trained our NMT models using tensor2tensor with its default hyper-parameters. Refer to \Sec{mtsettings} for further details on training configurations.

\begin{table}[t]
    \small
    \centering
    \begin{tabular}{|c|c|c|c|}
        \hline
        Dataset & Train & Dev & Test \\ \hline \hline
        ASPEC & 1.0M  & 1,790 & 1,812 \\ \hline
        TED & 223k  & 1,354 & 1,194 \\ \hline
    \end{tabular}
    \caption{Number of sentence pairs in each corpus.}
    \label{tab:aspected}
\medskip
    \begin{tabular}{|c|c|}  
        \hline
        Training schedule & BLEU\\ \hline \hline
        A & 4.1  \\ \hline
        T & 12.2  \\ \hline
        AT & 14.6 \\ \hline
        A {\ra} T& 13.9 \\ \hline
        A {\ra} AT & \textbf{15.0} \\ \hline
    \end{tabular}
    \caption{BLEU score for \task{Ja}{En} on TED test set.}
    \label{tab:initmt}
\end{table}

So far, we do not have a test set for the target domain, i.e., Coursera.  We therefore evaluated the performance of the MT systems with BLEU score \cite{papineni:02:b} on the TED test set.  \Tab{initmt} gives the results, where ``A,'' ``T,'' and ``AT'' stand for ASPEC, TED, and their balanced mixture, respectively and ``{\ra}'' means fine-tuning on the right-hand side data.  The model first pre-trained and then mixed fine-tuned, i.e., A{\ra}AT, gave the best result on the TED test set.  Thus, we used this model for sentence alignment.

\subsection{Creating Japanese--English Dataset} \label{sec:createjaen}

Finally, we extracted parallel sentences using the initial Japanese--English MT system to translate the Japanese sentences into English and the English embeddings available at the NLPL word embeddings repository\fn{\url{http://vectors.nlpl.eu/repository}, ID 40: Word2Vec Continuous Skipgram trained on English CoNLL2017 corpus. \href{http://vectors.nlpl.eu/repository/11/40.zip}{Download}} to compute sentence similarity. With the two parameters for constraining the DP algorithm, i.e., $\mathit{th}=0.92$ and $k=2$, we obtained a total of 43,549 pairs of sentences from 884 document pairs.

Then, following the procedure in \Sec{split}, we manually\fn{The checker is not a native English or Japanese speaker, but has the N1 certification (highest level) of the Japanese Language Proficiency Test and 99 points in TOEFL iBT.} created the test and development sets, taking the most reliable document pairs. We set 2,000 and 500 sentences as the target $\mathit{volume}$ for the test and development sets, respectively, and set $\mathit{ratio}=0.50$.
As shown in \Tab{testdevset},
a total of 2,779 sentence pairs drawn from 66 documents were manually judged in approximately 4 hours and about 8.4\% of them ((177+56)/2,779) were filtered out.

\begin{table}[t]
    \small
    \centering
    \begin{tabular}{|c|c|c|c|}
        \hline
        & \makecell{\# of \\ document pairs} &  \makecell{\# of \\ aligned lines} & \makecell{\# of \\deleted lines} \\ \hline \hline
        Test & 50  & 2,005 & 177  \\ \hline
        Dev & 16  & 541 &  56\\ \hline
        Train & 818 & 40,770 & - \\ \hline
    \end{tabular}
    \caption{Our Japanese--English Coursera parallel data.}
    \label{tab:testdevset}
\medskip
    \begin{tabular}{|c|c|c|}
        \hline
        \multirow{2}{*}{Dataset} &\multicolumn{1}{c|}{English} &\multicolumn{1}{c|}{Japanese} \\\cline{2-3}
        &Mean / Median / s.d. &Mean / Median / s.d. \\ \hline \hline
        ASPEC		& 25.4 / 23 / 11.4 & 27.5 / 20 / 12.0 \\ \hline
        TED		& 20.4 / 17 / 13.9 & 19.8 / 16 / 14.1 \\ \hline
        Coursera	& 21.1 / 19 / 11.1 & 22.2 / 20 / 11.8 \\ \hline
    \end{tabular}
    \caption{Statistics on the sentence length.}
    \label{tab:length-distribution}
\medskip
    \begin{tabular}{|c|c|c|c|}
        \hline
        \diagbox{LM}{Corpus} & ASPEC & TED & Coursera\\ \hline \hline
        ASPEC		& -1.147  & -3.013  & -2.926  \\ \hline
        TED		& -2.962  & -1.097  & -2.255  \\ \hline
        Coursera	& -2.658  & -2.335  & -0.760  \\ \hline
    \end{tabular}
    \caption{Per-token log-likelihood.}
    \label{tab:datasetsimilarity}
\end{table}

\subsection{Analysis} \label{sec:similarityanalysis}

We compared our Coursera dataset with the ASPEC and TED datasets regarding average sentence length and domain similarity.
\Tab{length-distribution} gives a summary of sentence length: the number of tokens segmented by Juman++ and NLTK for Japanese and English, respectively.  Coursera dataset is in between ASPEC and TED in its average sentence length, but relatively closer to TED than to ASPEC.

We also computed the similarity between datasets using language model (LM). First, we trained a 4-gram LM on the lower-cased version of English side of each training set. We then computed the per-token log-likelihood of these training sets with each of these LMs.  As shown in \Tab{datasetsimilarity}, three datasets are visibly distant to each other.  Nevertheless, TED seems relatively more exploitable than ASPEC for helping to translate Coursera datasets, presumably because they comprise spoken language unlike ASPEC.

\section{Japanese--English Lectures Translation}
We now describe how we can utilize the parallel corpus compiled as mentioned in the previous section for Japanese--English educational lectures translation.

\begin{table*}[t]
    \small
    \centering

\begin{tabular}{|c|ccccc|r|r|p{0mm}|c|cccc|r|r|}
\cline{1-8}\cline{10-16}
  ID &\multicolumn{5}{c|}{Training schedule} &\task{Ja}{En} &\task{En}{Ja}
&&ID &\multicolumn{4}{c|}{Training schedule} &\task{Ja}{En} &\task{En}{Ja}\\
\cline{1-8}\cline{10-16}

A1	&\bf A	&	&	&	&	&13.6 	&10.4 	&	&	&	&	&	&	&	&\\
A2	&\bf A	&\bgR{AT}	&	&	&	&25.6 	&13.5 	&	&B2	&\bf AT	&	&	&	&24.5 	&13.3\\
A3	&\bf A	&\bgR{AT}	&\bgR{AT}C	&	&	&\bf 27.5 	&18.0 	&	&B3	&\bf AT	&\bgR{ATC}	&	&	&26.8 	&17.0\\
A4	&\bf A	&\bgR{AT}	&\bgR{AT}C	&\bgB{TC}	&	&25.9 	&17.6 	&	&B4	&\bf AT	&\bgR{ATC}	&\bgB{TC}	&	&25.1 	&17.0\\
A5	&\bf A	&\bgR{AT}	&\bgR{AT}C	&\bgB{TC}	&\bgB{C}	&24.4 	&17.7 	&	&B5	&\bf AT	&\bgR{ATC}	&\bgB{TC}	&\bgB{C}	&23.8 	&17.7\\
A6	&\bf A	&\bgR{AT}	&\bgR{AT}C	&	&\bgB{C}	&24.7 	&\bf 18.5 	&	&B6	&\bf AT	&\bgR{ATC}	&	&\bgB{C}	&24.1 	&17.8\\
A7	&\bf A	&\bgR{AT}	&	&\bgK{TC}	&	&26.9 	&17.5 	&	&B7	&\bf AT	&	&\bgK{TC}	&	&26.4 	&17.2\\
A8	&\bf A	&\bgR{AT}	&	&\bgK{TC}	&\bgB{C}	&24.3 	&17.6 	&	&B8	&\bf AT	&	&\bgK{TC}	&\bgB{C}	&23.9 	&17.5\\
A9	&\bf A	&\bgR{AT}	&	&	&\bgK{C}	&23.8 	&17.2 	&	&B9	&\bf AT	&	&	&\bgK{C}	&22.9 	&17.7\\
A10	&\bf A	&	&\bgR{ATC}	&	&	&25.7 	&17.9 	&	&B10	&	&\bf ATC	&	&	&22.2 	&15.8\\
A11	&\bf A	&	&\bgR{ATC}	&\bgB{TC}	&	&25.2 	&17.4 	&	&B11	&	&\bf ATC	&\bgB{TC}	&	&22.0 	&15.4\\
A12	&\bf A	&	&\bgR{ATC}	&\bgB{TC}	&\bgB{C}	&24.3 	&17.5 	&	&B12	&	&\bf ATC	&\bgB{TC}	&\bgB{C}	&21.2 	&16.6\\
A13	&\bf A	&	&\bgR{ATC}	&	&\bgB{C}	&24.3 	&17.8 	&	&B13	&	&\bf ATC	&	&\bgB{C}	&21.2 	&16.5\\
A14	&\bf A	&	&	&\bgK{TC}	&	&25.4 	&17.6 	&	&B14	&	&	&\bf TC	&	&15.3 	&11.2\\
A15	&\bf A	&	&	&\bgK{TC}	&\bgB{C}	&23.8 	&17.1 	&	&B15	&	&	&\bf TC	&\bgB{C}	&16.1 	&12.2\\
A16	&\bf A	&	&	&	&\bgK{C}	&21.6 	&16.9 	&	&B16	&	&	&	&\bf C	&6.2 	&6.4\\\cline{1-8}\cline{10-16}
C3	&\bf A	&\bgR{AT}	&\bgB{T}	&	&	&24.0 	&12.2 	&	&D3	&\bf AT	&\bgB{T}	&	&	&23.2 	&12.2\\
C4	&\bf A	&\bgR{AT}	&\bgB{T}	&\bgR{TC}	&	&25.8 	&16.9 	&	&D4	&\bf AT	&\bgB{T}	&\bgR{TC}	&	&24.6 	&16.6\\
C5	&\bf A	&\bgR{AT}	&\bgB{T}	&\bgR{TC}	&\bgB{C}	&23.8 	&17.6 	&	&D5	&\bf AT	&\bgB{T}	&\bgR{TC}	&\bgB{C}	&22.3 	&17.0\\
C6	&\bf A	&\bgR{AT}	&\bgB{T}	&	&\bgK{C}	&23.4 	&17.3 	&	&D6	&\bf AT	&\bgB{T}	&	&\bgK{C}	&22.5 	&17.0\\
C10	&\bf A	&	&\bgK{T}	&	&	&23.9 	&12.2 	&	&D10	&	&\bf T	&	&	&17.5 	&8.9\\
C11	&\bf A	&	&\bgK{T}	&\bgR{TC}	&	&25.3 	&16.3 	&	&D11	&	&\bf T	&\bgR{TC}	&	&20.6 	&13.8\\
C12	&\bf A	&	&\bgK{T}	&\bgR{TC}	&\bgB{C}	&23.6 	&16.6 	&	&D12	&	&\bf T	&\bgR{TC}	&\bgB{C}	&19.8 	&14.4\\
C13	&\bf A	&	&\bgK{T}	&	&\bgK{C}	&22.7 	&16.9 	&	&D13	&	&\bf T	&	&\bgK{C}	&19.5 	&14.6\\\cline{1-8}\cline{10-16}
E14	&\bf A	&	&	&\bgR{AC}	&	&23.2 	&17.9 	&	&F14	&	&	&\bf AC	&	&16.2 	&13.6\\
E15	&\bf A	&	&	&\bgR{AC}	&\bgB{C}	&22.1 	&16.5 	&	&F15	&	&	&\bf AC	&\bgB{C}	&16.3 	&13.9\\\cline{1-8}\cline{10-16}

\end{tabular}


    \caption{BLEU scores for all the multistage training options examined in our experiment.  Models A1--A16 and B2--B16 represent all the 31 ($=2^{5}-1$) sub-paths of the A{\ra}AT{\ra}ATC{\ra}TC{\ra}C flow. Bold indicates the \textbf{initial training}, and red-, blue-, and grey-colored cells mean \colorbox{cR}{inflation}, \colorbox{cB}{deflation}, and \colorbox{cK}{replacement} of training data, respectively.}
    \label{tab:results}
\end{table*}

\subsection{Multistage Fine-Tuning}

Although NMT needs a large amount of parallel data to work well, its performance is very sensitive to the domain of the dataset and to the order in which datasets are included in the training. As such, it is common to divide training into multiple stages where each stage uses data from different domains to maximize the impact of the domain-specific training data. As we have larger parallel corpora from other domains, such as TED (0.2M pairs; non-educational spoken domain) and ASPEC (3.0M pairs; scientific domain), we can leverage domain adaptation techniques, such as fine-tuning and mixed fine-tuning \cite{raj17}. Furthermore, \citet{imankulova2019exploiting} and \citet{dabre-etal-2019-exploiting} showed that training in multiple stages where each stage contains different proportions of various types of training data leads to the best results. Following them, we decided to conduct an extensive experiment with multistage training with different proportions of training data from different domains at each stage.

\subsection{Datasets}

As in the previous section, we performed Juman++ and NLTK tokenization for Japanese and English, respectively.
Henceforth, we refer to the ASPEC training data of 1.0 million lines as ``A,'' the TED training data of 0.2 million lines as ``T,'' and the Coursera training data of 40k lines as ``C.'' When combining more than one dataset, we always oversample the smaller ones to match the size of the largest one. We denote the concatenated corpus by a concatenation of the letters representing them: e.g., AT for the mixture of ASPEC data with 5 times oversampled TED data, and ATC for the concatenation of ASPEC with 5 times oversampled TED data and 25 times oversampled Coursera data.

Following the observations in \Sec{similarityanalysis}, we decided to focus on the training schedule A{\ra}AT{\ra}ATC{\ra}TC{\ra}C, and thoroughly evaluated all of its sub-paths.
We also used T and AC for some contrastive experiments.

\subsection{Settings for MT}\label{sec:mtsettings}

We used the tensor2tensor framework \citep{tensor2tensor}\fn{\url{https://github.com/tensorflow/tensor2tensor}, version \href{https://github.com/tensorflow/tensor2tensor/releases/tag/v1.14.0}{1.14.0}.} with its default ``\emph{transformer\_base}'' setting, such as dropout=0.2, attention dropout=0.1, optimizer=adam with beta1=0.9, beta2=0.997.

We created a shared sub-word vocabulary for Japanese and English from ASPEC and TED training set using BPE \citep{sennrich-etal-2016-neural} with roughly 32k merge operations. This vocabulary was used for all experiments, even when a model is trained only on C.

In every experiment, we used eight Tesla V100 32GB GPUs with batch size of 4,096 sub-word tokens. We used early-stopping on approximate BLEU score computed on the development set: the training process stops when the score shows no gain larger than 0.1 for 10,000 steps. When fine-tune the model on a different dataset, we always resumed the training process from the last checkpoint in the previous stage.

In the decoding step, we always used the average of the last 10 checkpoints, and decoded the test sets with a beam size of 4 and a length penalty, $\alpha$, of 0.6 consistently across all the models.
The trained systems were evaluated with BLEU scores computed by sacreBLEU.\fn{\url{https://github.com/mjpost/sacreBLEU}}

\subsection{Results}

\Tab{results} summarizes the BLEU scores of all the MT systems trained up to five training stages.
The training schedule with all the stages, i.e., A{\ra}AT{\ra}ATC{\ra}TC{\ra}C (A5) did not achieve the best results for both translation directions. For the \task{Ja}{En} task, one of the intermediate models, A{\ra}AT{\ra}ATC (A3), gave the best BLEU score with more than 20 points gain over the model trained only on the in-domain parallel data (B16).  For the reverse direction, i.e., the \task{En}{Ja} task, the schedule A{\ra}AT{\ra}ATC{\ra}C (A6) achieved the best BLEU score with a 12.1 point BLEU gain.
In contrast, training a model directly in one stage on ATC (B10) gave significantly lower results than the multistage results.

Whenever new training data were introduced (marked \colorbox{cR}{red} in \Tab{results}), the BLEU scores were improved.\footnote{Compare the pairs (A1, A2), (A2, A3), (A1, A10), (B2, B3), (C3, C4), (C10, C11), (D3, D4), (D10, D11), and (A1, E14).}
This is mostly in line with the observations of \citet{raj17}; starting from out-of-domain and ending with a mixture of out-of- and in-domain data gives the best results for in-domain translation. As shown in \Tab{datasetsimilarity}, A is most dissimilar to C, and T is most similar to C. As such, it seems reasonable that gradually introducing the in-domain data by relying on related-domain data for intermediate training steps.
According to \citet{dabre-etal-2019-exploiting}, the final stage of fine-tuning on C should give the best translation quality.  However, in our setting, this holds true only for some cases in the \task{En}{Ja} task, suggesting the necessity of hyper-parameter tuning for fine-tuning on deflated training data (marked \colorbox{cB}{blue} in \Tab{results}).

This shows the importance of exhaustively exploring all settings, which confirmed and/or revealed the followings.
\begin{itemize}\itemsep=0mm
  \item Leveraging out-of-domain data through multistage training is invaluable.
  \item Gradually inflating the data starting from out-of-domain corpus and adding the in-domain corpus at the end should give the best possible translation quality.
\end{itemize}

\begin{table}[t]
    \small
    \centering
    \begin{tabular}{|l|l|c|c|}
    \hline
        Test set &Model & \task{Ja}{En} & \task{En}{Ja} \\ \hline \hline 
        ASPEC & A (A1 in \Tab{results}) & 29.8 & 41.5  \\ \hline
        TED & T (D10 in \Tab{results}) & 14.7 & 12.2 \\ \hline
        Coursera & C (B16 in \Tab{results}) &6.2 &6.4 \\ \hline
    \end{tabular}
    \caption{BLEU scores for different test sets.}
    \label{tab:unittest}
\end{table}

One peculiarity of our results is that the BLEU scores for the \task{Ja}{En} task were significantly higher than the \task{En}{Ja} task, which is reversal of general tendency for this language pair \citep{imamura-sumita-2018-multilingual,nakazawa-etal-2019-overview}, even though the BLEU scores in different languages are not directly comparable. \Tab{unittest} gives a comparison of three different translation tasks.
Upon manual investigation, we identified that the \task{En}{Ja} translations in TED and Coursera tasks tend to be much shorter than the reference translation, receiving around 0.7 brevity penalty. When we tuned the length penalty for decoding on the development set, we observed 0.5 to 1.0 point BLEU gains on the test set for the \task{En}{Ja} task, but this is not enough to flip the BLEU score tendencies.
Another possible reason is the nature of translationese \citep{rubino-etal-2016-information}.
Whereas ASPEC contains mainly Japanese-to-English translations, most talks in TED and Coursera datasets are English-to-Japanese translations.
Yet another reason is the difference between written (ASPEC) and spoken (TED and Coursera) languages.
We leave deeper exploration for the future.

\subsection{Iterative Refinement of Aligned Data}

Having obtained a better MT system than the initial one, we can iterate the whole process illustrated in \Fig{framework}, i.e., extracting the best possible parallel sentences using an MT system, and training a new MT system on the new parallel corpus, in order to maximize the quality of both the parallel corpus and the MT system.

To verify the impact of repeating an iteration, we took the best-performing \task{Ja}{En} MT system, i.e., A{\ra}AT{\ra}ATC (A3), and performed sentence alignment for the document pairs used as source for the training set, retaining test and development sets.
The re-aligned training data for C were used in the A{\ra}AT{\ra}ATC training schedule, where the models until A{\ra}AT were identical to those obtained in the first iteration, since they did not see C at all.

\begin{table}[t]
    \small
    \centering
    \begin{tabular}{|l|c|c|}
    \hline
        A{\ra}AT{\ra}ATC &
        \task{Ja}{En} & \task{En}{Ja} \\ \hline \hline 
        Iteration 1 (A3 in \Tab{results}) & \textbf{27.5} & \textbf{18.0}  \\ \hline
        Iteration 2 & 27.2 & 17.9 \\ \hline
    \end{tabular}
    \caption{BLEU scores in different iterations.}
    \label{tab:align-algo-2}
\end{table}

\Tab{align-algo-2} compares the BLEU scores achieved by the A{\ra}AT{\ra}ATC models in the first two iterations.
Unfortunately, we do not see any improvement in translation quality. We can speculate the following two reasons.
\begin{itemize}\itemsep=0mm
\item The dataset of approximately 40,000--45,000 sentences is too small to have any visible impact on translation quality.
\item The best possible sentence alignments for Coursera data were already found, owing to our algorithm, similarity measure, and/or the initial MT system trained only on ASPEC and TED.
\end{itemize}
Nevertheless, our observation does not necessarily hold for every language pair.
We thus encourage researchers to try iterative refinement of training data in their own experimental settings.

\subsection{Indirect Assessment of the Created In-Domain Data}

In this section, we evaluate the superiority of our sentence alignment method, presented in \Sec{alignment} (henceforth, MT+CS), over other methods, extrinsically, through MT performance.  The following two similarity measures were additionally implemented and tested.
\begin{description}\itemsep=0mm
    \item[Unsupervised:] Cosine similarity over the cross-lingual sentence embeddings, learned by an unsupervised method, called VecMap \citep{artetxe2017learning}.\fn{\url{https://github.com/artetxem/vecmap}}
    \item[MT+BLEU:] The metric in \Eq{bleu}, where the MT system was identical to the one used for our similarity metric, i.e., \Eq{emb}.
\end{description}

These two methods also rely on thresholding of the similarity scores of sentence pairs to get rid of potential noise. We set the threshold to a value such that the number of resulting sentence pairs is roughly the same as the number of pairs produced by our proposed method.

\begin{table}[t]
    \small
    \centering
    \begin{tabular}{|l|c|c|c|}
    \hline
        & \# of & \multicolumn{2}{c|}{BLEU} \\ &aligned lines &\task{Ja}{En} &\task{En}{Ja} \\ \hline \hline
        Unsupervised & 40,452 & 4.0 & 4.3  \\ \hline
        MT+BLEU & 42,672 & 2.8 & 3.4  \\ \hline
        MT+CS (B16 in \Tab{results}) & 40,770 & \textbf{6.2} & \textbf{6.4}  \\ \hline
    \end{tabular}
    \caption{BLEU scores achieved with only Coursera parallel data extracted by different similarity measures.}
    \label{tab:align-algo}
\end{table}

\Tab{align-algo} compares the number of extracted parallel sentences and the BLEU scores obtained by NMT systems trained only on the automatically aligned in-domain training data.  Whereas using BLEU as a measure of sentence similarity for alignment was bad, unsupervised cross-lingual embeddings gave more reliable similarity scores leading to better aligned sentences. Our MT+CS method which combines these two methods was able to give a parallel corpus, giving the highest BLEU score.

However, this does not completely justify the superiority of our similarity measure, because we have created the test and development sets relying on the MT+CS method.  This could introduce a bias in the resulting set toward this particular alignment method.  Another concern is the difficulty of the translations.  Even though we have obtained reasonably high BLEU scores in our experiment (\Tab{results}), due to heavy reliance on the word embeddings, the test and development sets may contain relatively easy sentence pairs in the sense that the sentence-level correspondences are easy to detect with such a simple method.  We plan to investigate these aspects in our future work.

\section{Conclusion and Future work}
In this paper, we proposed a framework to create a dataset for educational domain lectures translation. Specifically, we proposed a novel sentence similarity measure that combines machine translation and cosine similarity over sentence embeddings. Taking Japanese--English translation as a case study, we created a dataset of approximately 40,000, 500 and 2,000 lines of training, development, and test sets, with manual cleaning of the latter two sets to ensure that they can be used to reliably benchmark translation performance. We then utilized the automatically extracted parallel sentences to train an NMT system for Japanese--English lectures translation and show that multistage training in a domain adaptation framework leads to better translation models.

We will release our code used in our experiments for the sake of reproducibility. Given that the data crawled from Coursera is multilingually aligned at the document level, we plan to compile and provide a multilingual parallel corpus for lectures translation in the near future.

\section*{Acknowledgments}

This work was carried out when Haiyue Song was taking up an internship at NICT, Japan.
A part of this work was conducted under the program ``Research and Development of Enhanced Multilingual and Multipurpose Speech Translation System'' of the Ministry of Internal Affairs and Communications (MIC), Japan.

\section{References}
\bibliography{reference}
\bibliographystyle{lrec}

\end{document}